\documentclass[pdflatex,sn-mathphys-num]{sn-jnl}%
\usepackage{graphicx}%
\usepackage{multirow}%
\usepackage{amsmath,amssymb,amsfonts}%
\usepackage{amsthm}%
\usepackage{mathrsfs}%
\usepackage[title]{appendix}%
\usepackage{xcolor}%
\usepackage{textcomp}%
\usepackage{manyfoot}%
\usepackage{booktabs}%
\usepackage{algorithm}%
\usepackage{algorithmicx}%
\usepackage{algpseudocode}%
\usepackage{listings}%
\usepackage{geometry}
\usepackage[utf8]{inputenc}
\usepackage[T1]{fontenc}
\usepackage{upgreek}
\usepackage{xcolor}
\usepackage{subcaption}
\usepackage{caption}

\usepackage{hyperref}  % 用于创建超链接
\usepackage{nameref}   % 用于引用章节标题

\newcommand{\modelname}{ConcepPath}

\renewcommand{\abstractname}{}  % 取消 abstract 标题

\usepackage{xcolor}
\usepackage{ifthen}
\newcommand{\revisionone}{0} 
\newcommand{\showifred}[1]{\ifthenelse{\equal{\revisionone}{1}}{\textcolor{red}{#1}}{\textcolor{black}{#1}}}
\newcommand{\minorrevision}{0} 
\newcommand{\showifminor}[1]{\ifthenelse{\equal{\minorrevision}{1}}{\textcolor{red}{#1}}{\textcolor{black}{#1}}}

\theoremstyle{thmstyleone}%
\theoremstyle{thmstyletwo}%

\theoremstyle{thmstylethree}%

\begin{document}

\setcounter{secnumdepth}{0}  % 关闭章节编号

%\title[Article Title]{Towards Precise Histopathology Image Analysis: Aligning Knowledge Concepts to Whole Slide Images}

\title[Article Title]{Aligning Knowledge Concepts to Whole Slide Images for Precise Histopathology Image Analysis}


\author[1]{\fnm{Weiqin} \sur{Zhao}}\email{wqzhao98@connect.hku.hk}
\equalcont{These authors contributed equally to this work.}

\author[1]{\fnm{Ziyu} \sur{Guo}}\email{gzypro@connect.hku.hk}
\equalcont{These authors contributed equally to this work.}

\author[1]{\fnm{Yinshuang} \sur{Fan}}\email{fys0806@connect.hku.hk}

\author[2]{\fnm{Yuming} \sur{Jiang}}\email{yumjiang@wakehealth.edu}

\author*[3]{\fnm{Maximus} \sur{Yeung}}\email{mcfyeung@pathology.hku.hk}

\author*[1]{\fnm{Lequan} \sur{Yu}}\email{lqyu@hku.hk}

\affil[1]{\orgdiv{Department of Statistics and Actuarial Science}, \orgname{The University of Hong Kong}, \orgaddress{ \state{Hong Kong SAR}, \country{China}}}

\affil[2]{\orgdiv{School of Medicine}, \orgname{Wake Forest University}, \orgaddress{ \state{Winston-Salem, North Carolina}, \country{United States}}}

\affil[3]{\orgdiv{Department of Pathology}, \orgname{The University of Hong Kong}, \orgaddress{ \state{Hong Kong SAR}, \country{China}}}

% \affil[3]{\orgdiv{Department}, \orgname{Organization}, \orgaddress{\street{Street}, \city{City}, \postcode{610101}, \state{State}, \country{Country}}}


\abstract{
Due to the large size and lack of fine-grained annotation, Whole Slide Images (WSIs) analysis is commonly approached as a Multiple Instance Learning (MIL) problem.
However, previous studies only learn from training data, posing a stark contrast to how human clinicians teach each other and reason about histopathologic entities and factors.
Here we present a novel knowledge concept-based MIL framework, named \textbf{\modelname} to fill this gap.
Specifically, \modelname\ utilizes GPT-4 to induce reliable disease-specific human expert concepts from medical literature, and incorporate them with a group of purely learnable concepts to extract complementary knowledge from training data.
In \modelname, WSIs are aligned to these linguistic knowledge concepts by 
\protect\showifred{utilizing pathology vision-language model as the basic building component.}
In the application of lung cancer subtyping, breast cancer HER2 scoring, and gastric cancer immunotherapy-sensitive subtyping task, \modelname\ significantly outperformed previous SOTA methods which lack the guidance of human expert knowledge.
}

%%================================%%
%% Sample for structured abstract %%
%%================================%%

\maketitle

\section{Introduction}
\begin{figure}[!ht]
    \centering
    \includegraphics[width=0.98\textwidth]{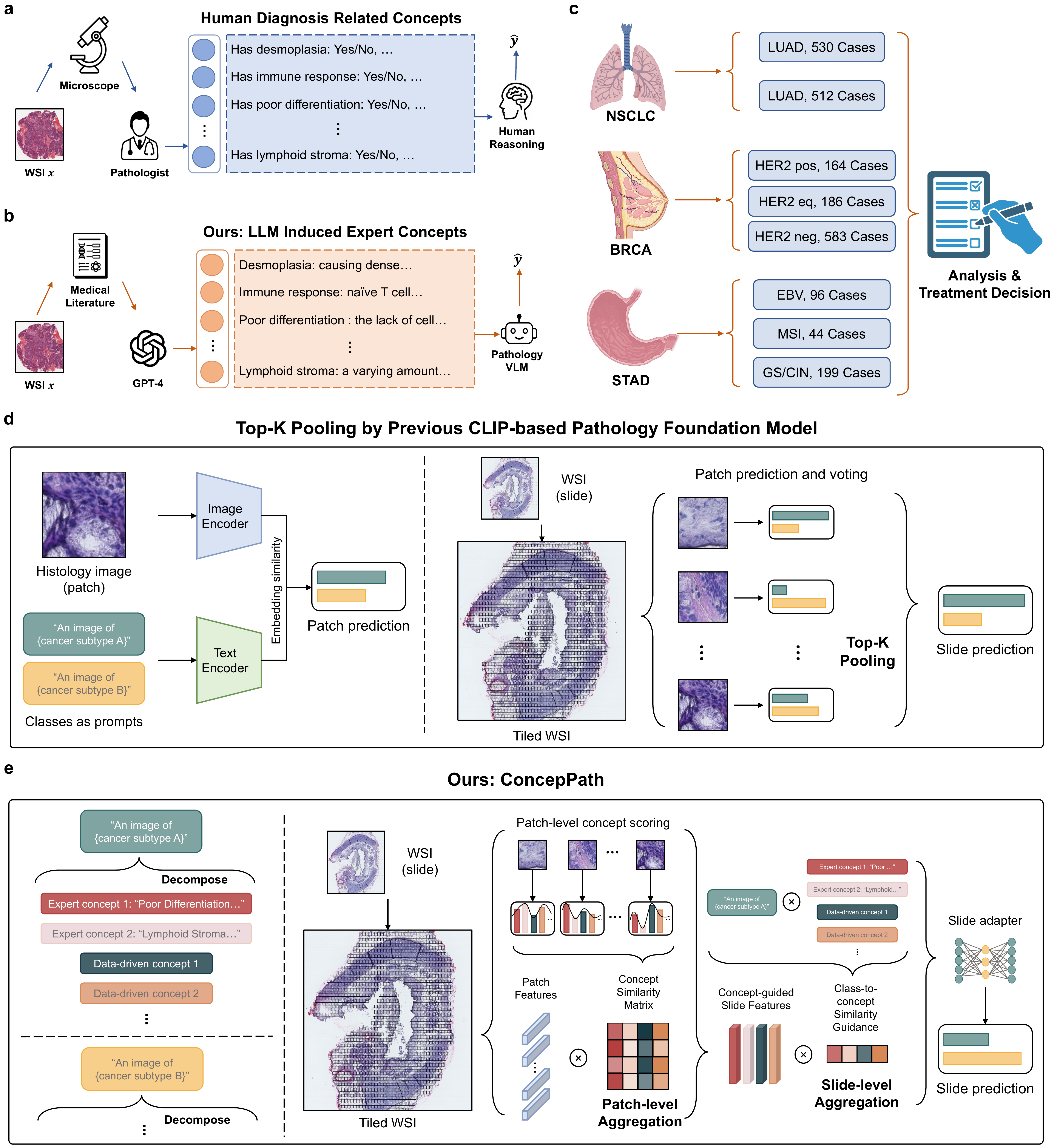}
    
    \caption{
            \textbf{Overview of \modelname\ framework.}
    \textbf{a,} In real clinical processes, pathologists apply their expert knowledge to reason about histopathologic entities and factors to make a diagnosis. 
    \textbf{b,} \modelname\ utilizes a large language model like GPT-4 to induce expert concepts related to diagnosis from medical literature and integrate this knowledge into an automated WSI analysis pipeline through the \protect\showifred{CLIP-based pathology vision-language foundation model.}
    \protect \showifred{\textbf{c,} Dataset Characteristics of the evaluated tasks.}
    % \protect \showifred{\textbf{d,}}
    % Overview of \modelname. Top Left: \modelname\ obtains the embeddings of the instance-level expert concepts and the complementary data-driven concepts which are optimized with the training data during the learning process. Top Right: \modelname\ obtains the embeddings of the bag-level expert class prompts. Bottom: \modelname\ aggregates instance features of the WSI into an overall bag representation through a hierarchical two-stage aggregation paradigm, guided by instance-level concepts and correlations between bag-level class prompts and instance-level concepts.
    % }
    \protect\showifminor{
        \textbf{d,} Left: An illustration of how the CLIP-based pathology foundation model performs patch prediction with class prompts. Right: The pipeline of how the previous CLIP-based pathology foundation model performs slide-level classification via a top-k pooling of patch predictions.
        \textbf{e,} Left: An illustration of how \modelname\ decomposes a specific complex WSI analysis task into multiple subtasks of scoring patch-level concepts/attributes. Right: The pipeline of how \modelname\ conducts slide-level classification. Unlike the previous CLIP-based pathology foundation models' mechanism, \modelname\ leverages human prior knowledge and fully exploits the power of the CLIP-based pathology foundation model by scoring a group of expert concepts induced by GPT-4 from related medical literature, and extracting complementary knowledge from training data via scoring a group of learnable data-driven concepts. The final prediction is produced with a two-stage aggregation mechanism with the above concepts.
    }
    }
    
    \label{fig:overview}
\end{figure}

The analysis of histopathology images is crucial in modern medicine, particularly for cancer diagnosis and prognosis, where it serves as the gold standard. 
However, analyzing histopathology images is time-consuming and labor-intensive for pathologists.
Digitalizing histopathology images into high-resolution whole slide images (WSIs) has ushered in a new era for computer-aided analysis~\cite{wang2019weakly, wang2019pathology, coudray2018classification}. 
Owing to their enormous size (e.g., 150,000 x 150,000) and the lack of fine-grained annotations, WSI analysis is typically formulated as a Multiple Instance Learning (MIL) problem, which enables weakly supervised learning from slide-level labels.
MIL-based methods typically begin by extracting the feature embeddings of image patches using a pre-trained network~\cite{he2016deep, riasatian2021fine, chen2022scaling, wang2022transformer}. 
The feature embeddings are then fed into an aggregation network to generate slide-level predictions. 
Numerous research efforts have focused on efficiently aggregating information, including using attention-based weights~\cite{ilse2018attention} and leveraging spatial context information~\cite{hou2022h2, guan2022node, chan2023histopathology}. 
% \ylq{shorten the above sentences.}
%
However, most current approaches in computational histopathology learn solely from image data, contrasting with how humans teach and reason about histopathologic entities and factors, as illustrated in Figure~\ref{fig:overview}a. 
Although a recent innovative study~\cite{qu2023rise} explores the use of language priors in few-shot weakly supervised learning for WSI analysis, it suffers from unreliable language prior generation and unsatisfactory performance under full training setups, which limits its wide application in precise WSI analysis.
Thus, incorporating valuable expert knowledge for precise WSI analysis remains an unsolved yet critical challenge. 
With the rapid development of multi-modal learning, there has been a surge of studies on 
\protect\showifred{CLIP-based pathology vision language models~\cite{gamper2021multiple, lu2023visual, radford2021learning, huang2023leveraging, ikezogwo2023quilt, lu2023towards}. }
Following the principle of Contrastive Language-Image Pre-training (CLIP)~\cite{radford2021learning}, these models learn well-aligned representation spaces between histopathology images and text description pairs collected from medical textbooks, scientific papers, public forums, and educational videos.
One major benefit is that natural language descriptions can provide more expressive, dense, and interconnected representations beyond the scope of a single categorical label, linking diverse features of histopathology sub-patch structures~\cite{gamper2021multiple, lu2023towards}. 
Remarkable achievements have been made in transferring the above pathology vision-language models to a wide range of downstream tasks, including patch-level histopathology image classification, segmentation, captioning, and retrieval~\cite{lu2023towards}.
Meanwhile, large language models (LLMs) have shown great potential in performing logic, analogy, causal reasoning, extrapolation, and evidence evaluation for medical and scientific applications~\cite{truhn2023large}. 
Researchers have found that when treated as reasoning machines or inference engines rather than knowledge databases, LLMs are less likely to generate false statements that do not reflect scientific facts~\cite{sanderson2023gpt}.
These breakthroughs provide an opportunity to extract and incorporate human expert knowledge into the WSI analysis.
%-------------------------------------------------------------------------

In this paper, we present \textbf{\modelname}, a concept-based framework designed to make decisions by jointly leveraging the complementary human expert prior knowledge and data-driven concepts.
The main idea of \modelname\ is illustrated in Figure~\ref{fig:overview}b and     \protect\showifminor{Figure~\ref{fig:overview}e}. 
%-------------------------------------------------------------------------
%
First, large language models such as GPT-4 are applied to induce reliable \showifred{instance-level human expert concepts and bag-level expert class prompts} from medical literature highly relevant to the given diagnostic task, as shown in Figure~\ref{fig:overview}b \showifred{and Supplementary Figure~2}.
It is important to note that, instead of treating GPT-4 as a knowledge database and directly querying expert concepts, \modelname\ utilize GPT-4 as a reasoning machine to induce human expert knowledge concepts \showifred{and class prompts} from medical textbooks and academic papers. 
This strategy leads to more reliable expert concept \showifred{and class prompts} generation.
On the other hand, considering that extracted expert concepts may be insufficient to fully describe a disease's complexity~\cite{echle2021deep, mandair2023biological},  \modelname\ also learns complementary data-driven instance-level concepts from the training data with a group of purely learnable prompt representations, as shown in     \protect\showifminor{Figure~\ref{fig:overview}e}. 
These learned concepts serve as a complement to expert concepts and play a crucial role, especially for complex and inadequately researched diagnostic tasks.

To transfer the knowledge contained in the concepts into WSI analysis, \modelname\ utilized a two-stage concept-guided hierarchical feature aggregation paradigm.
With both concepts, \showifred{class prompts,} and instance features embedded into the well-aligned representation space via \showifred{the CLIP-based pathology vision-language model as the basic building component}, \modelname\ first aggregate instance features into concept-specific bag-level features under the guidance of instance-level concepts and then further aggregate the concept-specific bag-level features into the overall bag representation according to the correlations between instance-level concepts and \showifred{bag-level expert class prompts}.
Finally, \modelname\ feeds the overall bag representation and bag-level concept embedding to slide-adapters and calculates similarities between the adapted bag representations and \showifred{bag-level expert class prompts} embeddings for final prediction.
We validated the effectiveness of \modelname\ on five tasks (Figure~\ref{fig:overview}c): (1) lung cancer subtyping, (2) breast cancer HER2 scoring, and (3) gastric cancer immunotherapy-sensitive subtyping (including 3 binary classification tasks).
\modelname\ outperformed previous state-of-the-art methods on all tasks in Figure~\ref{fig:performance}a, which shows the prominence of utilizing human expert knowledge effectively. 
Particularly noteworthy is the nearly $7\%$ improvement that \modelname\ achieved on classifying  Epstein–Barr virus(EBV)-positive for gastric cancer cases, demonstrating its potential as economical and less time-consuming alternatives for stratifying patients who respond to immune checkpoint inhibitor therapy.

\label{sec:intro}
%

%-------------------------------------------------------------------------

\section{Results}\label{results_sec}
%-----------------------------------------------------------------------------------------------------------------------------------------------------------------------------------------------------------------------------------------------------------------------------------------------------------------------------------------------------------------------------------------------------------------------------------------------------------------------------------------------------------------------------------------

% \section{Experiments}
% \label{Experiments}

\subsection{Dataset Characteristics for Tumor Diagnosis}
We evaluate \modelname\ on three public datasets (NSCLC, STAD and BRCA) from The Cancer Genome Atlas (TCGA) repository. 
The first dataset is \textbf{NSCLC}, the lung cancer project containing 1,042 cases.
For the tumor subtyping task on this dataset, there are 530 cases diagnosed as lung adenocarcinoma (LUAD) and 512 cases diagnosed as lung squamous cell carcinoma (LUSC).
The second dataset is \textbf{BRCA}, the breast cancer project containing 933 cases.
For the HER2 scoring task on this dataset, there are 164 cases diagnosed as positive, 186 cases diagnosed as equivocal and 583 cases diagnosed as negative.
Human epidermal growth factor receptor 2 (HER2) plays a crucial role as both a prognostic and predictive marker, being over-expressed in approximately $15-20\%$ of breast cancer cases.
Assessing HER2 status is vital for guiding clinical treatment choices and prognostic assessments. 
The evaluation of HER2 status is performed through transcriptomics or immunohistochemistry (IHC) methods, including in-situ hybridization (ISH), which adds extra costs and tissue demands. 
Furthermore, this process is subject to variability in analysis, especially due to potential biases in manual scoring observations~\cite{wolff2013american, lu2022slidegraph+}.
The last dataset is \textbf{STAD}, the gastric cancer project containing 268 cases.
For the immunotherapy-sensitive subtyping task on this dataset, there are 26 cases diagnosed as Epstein–Barr virus(EBV)-positive, 44 cases diagnosed as Microsatellite Instability (MSI), and 199 cases diagnosed as Genomically Stable (GS) and Chromosomal Instable (CIN).
EBV and MSI tumors have been reported to be highly responsive to Immune checkpoint inhibitor (ICI) therapy, which is widely used but effective only in a subset of gastric cancers~\cite{kelly2017immunotherapy}.
However, the high costs of required diagnostic methods like immunohistochemistry and polymerase chain reaction limit the practical application of this molecular classification in treatment decisions~\cite{hinata2021detecting}.
EBV-associated and MSI gastric cancers are characterized by distinct histological traits. EBV-positive tumors often display significant lymphocyte infiltration within both the neoplastic epithelium and the stroma, frequently referred to as lymphoepithelioma-like carcinoma or gastric carcinoma with lymphoid stroma~\cite{fukayama2020thirty}. 
Similarly, the MSI subtype typically shows extensive lymphocytic infiltration, predominantly featuring intestinal-type histology and expansive growth patterns~\cite{grogg2003lymphocyte, arai2013frequent}.
Note that in this study, we follow the setting in \cite{hinata2021detecting} to formulate three binary classification tasks: EBV vs. Others, MSI vs. Others, and EBV+MSI vs. Others.
Given the biases inherent in manual evaluations and the additional expenses associated with these assessments, particularly for the latter tasks corresponding to guiding clinical treatment choices, conducting precise analysis directly from routine H\&E-stained tissue sections through deep learning techniques is of significant clinical and scientific interest. 
On these tasks, We utilized patient-level five-fold cross-validation for all experiments evaluating \modelname\ and other methods and reported the results of all models in the form of mean value.
For evaluation metrics, the area under the curve (AUC) of receiver operating characteristic, and the accuracy (ACC) were adopted.
\showifred{\modelname\ could utilize different CLIP-based pathology vision-language foundation models as its basic component. As QuiltNet~\cite{ikezogwo2023quilt} and CONCH~\cite{lu2023visual} lead to better performance in our experiments, in the following sections, we report the results of using QuiltNet as the default basic component of \modelname\ and the feature extractor of other baselines in the following sections, and report the results of using CONCH in the Supplementary.}
%
%-------------------------------------------------------------------------

%-------------------------------------------------------------------------

\subsection{\modelname\ Helps in Treatment Decision}
We evaluated \modelname\ against seven state-of-the-art (SOTA) methodologies: (1) ABMIL~\cite{ilse2018attention}, (2) DeepAttnMISL~\cite{yao2020}, (3) CLAM-SB~\cite{lu2021data}, (4) GTP~\cite{zheng2021deep}, (5) TransMIL~\cite{shao2021transmil}, (6) HIPT~\cite{chen2022scaling}, and (7) TOP~\cite{qu2023rise}. For TOP~\cite{qu2023rise}, 
\showifred{due to the current incomplete official implementation released by the authors, we run both the author's current implementation and our re-implementation according to their paper, and select the higher performance as the final result of TOP.}
As illustrated in Figure~\ref{fig:performance}a, \modelname\ surpasses the aforementioned methods in both AUC and ACC across all assessed tasks. Specifically, a significant performance leap for \modelname\ is observed in the BRCA and STAD datasets. For example, on the BRCA dataset, \modelname\ registers a 5.66$\%$ increase in the F1-score over the leading baseline. Similarly, on the STAD dataset, improvements of 6.23$\%$ in EBV vs. Others AUC, 1.71$\%$ in MSI vs. Others AUC, and 1.35$\%$ in EBV+MSI vs. Others AUC were noted in comparison to the best-performing baseline.
These datasets involved complex histological analysis tasks such as HER2 scoring and immunotherapy-sensitive subtyping, necessitating the recognition of intricate histological features and molecular tissue characteristics. This leads us to hypothesize that for more intricate WSI analysis challenges, the fusion of prior domain knowledge with the discovery of new, diagnosis-related concepts is crucial, significantly more so than for simpler tumor/normal classification or tumor subtyping tasks.
Overall, \modelname\ enhances WSI analysis capabilities, especially in tumor subtyping and immune response assessment, potentially aiding in treatment decision-making processes.
Regarding TOP~\cite{qu2023rise}, designed to leverage language priors in few-shot weakly supervised learning for WSI analysis, it demonstrated unsatisfactory results in full training settings. This could be due to the unreliable generation of language priors, a misalignment between histopathology images and prior knowledge text, and a lack of new knowledge acquisition from the training data,
\protect \showifred{which is detailed in Section~\nameref{sec:baseline}.}
\protect \showifred{
In addition, we report the experimental results when using CONCH as the basic component of \modelname\ and feature extractor of other baselines in Supplementary Figure~5. 
Although HIPT achieved improved performance as CONCH was trained on a larger number of histology images, we noticed that \modelname\ still outperforms other baselines among most tasks and shows as a more robust method compared with others.
}
\subsection{Comparison of Different Expert Concept Extraction Strategies.}
The incorporation of human expert knowledge is of great significance to \modelname, the performance will be influenced if such prior is not imported accurately.
We investigate the impact of different expert concept generation strategies and show the mean results in Figure~\ref{fig:performance}d. 
The ``Generated concepts” refers to the strategy used in previous works~\cite{yang2023language, yan2023learning, qu2023rise}, which involves directly querying GPT-4 for relevant concepts without providing any expert materials. 
In contrast, ``Induced concepts” represent our proposed strategy, which entails asking GPT-4 to induce relevant concepts from medical literature related to the target diagnostic task.
As shown in Figure~\ref{fig:performance}d and Supplementary Figure~4a, the ``Induced concepts” achieve better performance across all metrics, particularly for the more challenging immunotherapy-sensitive subtyping tasks on the STAD dataset. 
This highlights the importance of inducing concepts from professional materials for complex WSI analysis tasks.
\protect \showifred{
In addition, we report the experimental results when using CONCH as the basic component of \modelname. 
As shown in Supplementary Figure~6, we could obtain the same conclusion as above.
}
In Figure~\ref{fig:performance}b and Figure~\ref{fig:performance}c, we also provide one example of a misleading concept generated by directly querying GPT-4, which our induction-based strategy successfully rectifies.
For the gastric immunotherapy-sensitive subtyping task, the expert concept ``signet-ring cells” is more commonly linked to the Genomically Stable (GS) subtype instead of Epstein–Barr virus (EBV) positive subtype~\cite{Ratti2018}.
However, as as illustrated in Figure~\ref{fig:performance}b the ``Generated concepts” strategy attributes the expert concept ``signet-ring cells” to the EBV positive subtype instead of the GS subtype, which may cause confusion for the model.
In contrast, as demonstrated in Figure~\ref{fig:performance}c, under the guidance of relevant medical literature, our proposed ``Induced concepts” strategy accurately attributes expert concept ``signet-ring cells” to GS subtype, which provides reliable prior expert knowledge to our framework.

\subsection{Effectiveness of Data-driven Concept Learning.}
The integration of learnable concepts is also a key component of \modelname. 
We investigate its impact on NSCLC lung cancer subtyping, BRCA HER2 scoring, and STAD EBV vs. others classification tasks. Results are shown in Figure~\ref{fig:ablation}a. 
The x-axis in Figure~\ref{fig:ablation}a represents the number of learned concepts used for each class, where 0 refers to the only use of GPT-4 induced expert concepts in our framework.
From the results, we observe a performance increase of \textbf{1.04\%} in AUC for NSCLC, \textbf{1.16\%} in AUC for BRCA and \textbf{3.96\%} in AUC for EBV vs. Others upon integrating new concept discovery, demonstrating the effectiveness of complementing human expert knowledge with learned knowledge. 
Furthermore, we noticed that for more challenging and inadequately researched diagnostic tasks, a greater number of learned concepts are required.
For instance, NSCLC achieved its best performance with $4$ learned concepts per class, while the EBV vs. Others comparison reached its peak performance with $8$ learned concepts per class.
Furthermore, we identified a performance decline when the number of learned concepts was large (.ie $12$) on both datasets. 
This observation suggests a potential trade-off between prior expert knowledge and learned knowledge. 
If the number of learned data-driven concepts is excessive, the impact of prior expert knowledge may be diminished, potentially resulting in overfitting the training data.
\protect \showifred{
We noticed that such a trade-off still exists when using CONCH as the basic component of \modelname, as shown in Supplementary Figure~7.
}
\protect\showifminor{
Particularly, we also investigated the independent contributions of the expert concept and the data-driven concept in \modelname\ (Supplementary Figure~11). 
We noticed that while removing the data-driven concepts will cause a bigger average performance drop, either ignoring the human expert prior or the knowledge in the training data would generally cause an obvious performance drop for most tasks on both QuiltNet-based and CONCH-based \modelname.
These results illustrate that the success of \modelname\ lies in incorporating the complementary human expert prior and data-driven knowledge in automated WSI analysis.
}

\subsection{Effectiveness of Bag-level Concept Guidance.}
We also investigate the impact of the bag concept-guided aggregation in \modelname.
As shown in Figure~\ref{fig:ablation}b and Supplementary Figure~4b, ``w/o Bag-level guidance” bars denote directly averaging the concept-specific bag-level features into the overall bag representation without considering the correlations between the instance-level concepts and the \protect \showifred{bag-level class prompts}.
The model performed better with Bag-level guidance, for instance, we observe a performance drop of \textbf{2.82\%} in AUC for EBV+MSI vs. Others and \textbf{4.11\%} in AUC for MSI vs. Others when ignoring the relationship between the instance-level concepts and the \protect \showifred{bag-level class prompts}.
This demonstrates the importance of the second-stage bag-level concept-guided aggregation in our framework.
\protect \showifred{
The experimental results when using CONCH as the basic component of \modelname, the overall performance among five tasks gain notable improvement with this second-stage bag-level concept-guided aggregation, especially for the more reflective metric AUC (Supplementary Figure~8).
}
%
% \textcolor{red}{Supp ACC Figure}

%

\subsection{Effectiveness of Slide-Adapters.}
The slide-adapters are proposed to learn new features and blend them with the original features of the overall bag representation and bag-level concept embedding. 
We also explore their effectiveness in Figure~\ref{fig:ablation}b and Supplementary Figure~4b.
Particularly, we notice an obvious performance drop in AUC for EBV vs. Others, which may indicate the potential limits of the knowledge within the existing human medical research papers and the pathology vision-language model with respect to this challenging and inadequately researched diagnostic task.
\protect \showifred{
The experimental results when using CONCH as the basic component of \modelname, the overall performance among five tasks would be improved by the slide-adapters, especially for the more reflective metric AUC (Supplementary Figure~8).
}

\subsection{Comparison of Different Vision-Language Models.}
The alignment of histopathology images with textual expert concepts is of great significance in our framework. 
\protect \showifred{
We also compare the efficiency of using different CLIP-based vision-language models as the basic building component in \modelname\ to align the concepts with histopathology images. 
}
Results are shown in Figure~\ref{fig:ablation}c and and Supplementary Figure~5c, where 
CLIP~\cite{radford2021learning} is trained on a variety of image-text pairs from the internet,
\protect \showifred{
PLIP~\cite{huang2023leveraging} and PathClip~\cite{zheng2024benchmarking} are trained on over 200K histopathology image-text pairs, 
and QuiltNet~\cite{ikezogwo2023quilt} and CONCH~\cite{lu2023visual} are trained on over 1 million histopathology image-text pairs.
}
Comparing the performances between CLIP and PLIP, an obvious performance increase can be observed by using pathology vision-language models.
Moreover, the additional performance increase brought by \protect \showifred{QuiltNet~\cite{ikezogwo2023quilt} and CONCH~\cite{lu2023visual}} further demonstrates that our framework could benefit from more accurate alignment, and thus more efficiently incorporating concept knowledge into histopathology images.
%
% \textcolor{red}{Supp ACC Figure}

%-------------------------------------------------------------------------

\subsection{Visualization and Post-hoc Interpretation}

Model interpretation is crucial for medical applications. \modelname\ offers post-hoc interpretation by visualizing the similarity scores between instance-level features and instance-level concepts as similarity maps on the slide, providing multi-dimensional reference information compared to previous attention map-based approaches.
% \para{Instance-level Concept Similarity Map.}
%Our framework offers instance-level concept similarity maps for model interpretation and diagnosis. 
%
Some visualization examples are presented in Figure~\ref{supp_sim}, which displays four distinct expert instance-level concept similarity maps for four accurately classified lung squamous cell carcinoma (LUSC) and lung adenocarcinoma (LUAD) slides. 
Additionally, we include the attention maps of CLAM~\cite{lu2021data} for a comprehensive comparison.
\protect\showifminor{
Evaluated by our expert pathologist collaborator, we note that the clinical relevance of this work lies in enabling pathologists to understand the rationale behind the model’s predictions for a given WSI. The heatmaps for various expert concepts pinpoint the exact regions within WSIs that contribute to specific predictions, providing a clear and interpretable basis for the model’s decisions. This is well-illustrated in Figure~\ref{supp_sim} and Supplementary Figure~12.
For instance, our pathologist collaborator identified that concepts such as "lepidic," "papillary," "glandular," "micropapillary," and "solid growth" are closely associated with LUAD, while "keratinization," "cell morphology," "nuclear changes," and "high mitoses" are characteristic of LUSC. These observations align with established medical knowledge, demonstrating the interpretability and clinical validity of the model. Notably, these expert concepts exhibit high activity in tumor regions, consistent with domain expertise.
Furthermore, similarity maps generated by \modelname\ provide a more precise focus on tumor regions compared to CLAM attention maps. For example, in the slide from the first row, the green box area includes additional tumor foci that are overlooked by CLAM. Similarly, benign mucous glands intermixed with inflammatory cells activate medical concepts such as inflammatory cells, associated lymphocytes, and immune cell clusters, reflecting plausible biological phenomena (Supplementary Figure~12).
Lastly, we observed minor variations in WSI regions across different expert concepts. These variations could serve as valuable supplementary references, offering multi-dimensional insights for pathologists during diagnosis.
}

\section{Discussion}\label{discussion_sec}

Despite the rapid advancement in computational pathology, the integration of valuable human expert knowledge into automated AI-assisted diagnosis remains a significant yet unresolved challenge. 
The advent of \protect\showifred{CLIP-based pathology vision-language foundation models} and large language models (LLMs) presents a promising avenue by aligning histopathology images with human linguistic priors for more efficient WSI analysis. 
However, direct queries to LLMs may yield inaccurate statements not grounded in scientific fact. 
Furthermore, the complexity of diseases may surpass the existing human expert knowledge, necessitating the discovery of complementary knowledge hidden within training data.
To address these issues, we introduce \modelname\ in this study, a novel framework that explicitly incorporates reliable prior expert knowledge and learns complementary concepts from training data for precise WSI analysis. 
To circumvent inaccuracies inherent in LLMs, \modelname\ employ GPT-4 as a reasoning engine to derive reliable instance-level expert concepts and \protect\showifred{bag-level expert class prompts} from medical literature \protect\showifred{related to the target diagnostic task}. 
Additionally, \modelname\ explores complementary data-driven instance-level concepts from the training data using a set of learnable prompt representations. 
Following the Multiple Instance Learning (MIL) paradigm, \modelname\ utilized a concept knowledge-guided two-stage hierarchical feature aggregation process for efficient bag-level WSI representation. 
Importantly, \modelname\ integrates slide-adapters prior to the final prediction to address the domain shift between the pathology vision-language model's training data and downstream WSI analysis tasks.
\protect\showifred{
To avoid confusion and distinguish \modelname\ from the CLIP-based pathology foundation models, we would like to emphasize that the objectives of the CLIP-based pathology foundation models and \modelname\ are in fact quite different: 
the CLIP-based pathology foundation models aim to align path-level histology images with their corresponding text description, while \modelname\ aims to provide precise and interpretable slide-level analysis via leveraging human expert prior knowledge and complementary knowledge extracted from training data. 
The CLIP-based pathology foundation models mainly serve as a basic building component in \modelname\ in this study.
We provide more detailed comparison and illustration of \modelname's advantage in Section~\nameref{Difference from CLIP-based Models}.
}
The proposed \modelname\ pioneers the incorporation of human expert knowledge for efficient and accurate histopathology analysis, a topic of significant clinical importance yet largely unexplored in existing research. 
The comprehensive experimental results in this study affirm the superiority of \modelname\ over state-of-the-art methods and demonstrate the efficacy of the proposed components, offering new perspectives on leveraging human expert knowledge, LLMs, \protect\showifred{CLIP-based pathology vision-language foundation models}, and training data for precise automated WSI analysis.

However, \modelname\ has limitations. Although it surpasses other state-of-the-art methods, it relies on additional supervised information, namely, expert knowledge induced by GPT-4, which may be insufficient or biased for rare or novel cancer types. 
Furthermore, the potential for divergent expert opinions on controversial topics underscores the challenge of acquiring accurate expert information — an issue we aim to address in future work. 
Additionally, while \modelname\ proposes learning data-driven concepts from training data, interpreting these concepts to enhance model understanding and facilitate medical discovery remains crucial. 
The current approach employs concept similarity maps for interpretation, but further collaboration with expert pathologists is necessary to decode the semantic information underlying different learned concepts. 
On the other hand, due to computational constraints, the text and image encoders remain frozen during training, despite the domain shift between encoder training data and downstream task data. 
Exploring resource-efficient strategies for fine-tuning both encoders on downstream tasks to mitigate this domain shift will also facilitate further research.
\protect \showifred{
Lastly, as PathChat~\cite{lu2024multimodal}, a multimodal generative vision-language AI assistant for human pathology, emerged as a concurrent work of \modelname, it would be promising if we could adopt PatChat as a basic building component in \modelname.
Specifically, PathChat could generate a response to both histology image and text input as a chatbot like GPT4V, which could provide a more flexible linkage between the histology image and human language.
However, as the PathChat model is not publicly available yet, we leave integrating PathChat in \modelname\ to explore whether it can improve \modelname's capability as our future work.
}

For other future works, we will aim to refine the \modelname\ framework, specifically addressing the domain shift between the encoders' training data and the downstream task data. 
This enhancement will focus on adapting the model to better generalize across different data sets, thereby improving its robustness and accuracy in diverse clinical scenarios. Furthermore, we plan to quantify and mitigate the challenges associated with obtaining precise expert information tailored to specific WSI analysis tasks, enhancing the reliability of the expert knowledge integrated into the model.
Additionally, we will explore the integration of graph representations into the \modelname\ framework to enrich the analysis of WSIs. By capturing the intricate spatial relationships and contextual details inherent in tissue structures, graph representations can provide a more nuanced understanding of the histopathological features. This advancement is anticipated to offer deeper insights into the tissue architecture, potentially unveiling new biomarkers and improving diagnostic accuracy.

\section{Methods}\label{methods_sec}

\subsection{\modelname\ Overview}
% \para{Overview.} 
%
Figure~\ref{fig:overview}b and \protect\showifminor{Figure~\ref{fig:overview}e} present our proposed \modelname\ framework. 
Specifically, as depicted in Figure~\ref{fig:overview}b, \modelname\ first utilizes the large language model, like GPT-4, to induce reliable disease-specific instance-level expert concept and \protect\showifred{bag-level expert class prompts} from medical literature. 
On the other hand, to complement the extracted expert knowledge, \modelname\ employs a set of purely learnable instance-level concepts for complementary data-driven instance-level concepts learned from the training data.
Next, \modelname\ aligns the histopathology patches and the concepts by leveraging CLIP-based pathology vision-language foundation models. 
Subsequently, the instance features are aggregated into the overall bag representation using a two-stage hierarchical aggregation paradigm, guided by the instance-level concept and the correlations between instance-level expert concepts and \protect\showifred{bag-level expert class prompts}.
Afterward, \modelname\ feeds the overall bag representation and \protect\showifred{bag-level expert class prompt} embeddings to slide-adapters, which serve as an additional bottleneck layer to perform residual-style feature blending with the original features.
Finally, the predictions are calculated based on the similarities between the adapted bag representations and \protect\showifred{bag-level expert class prompt} embeddings.
For simplicity, we elaborate \modelname\ with a binary classification WSI analysis task in the formulas in the following sections, identifying class $A$ and class $B$.
Note that \modelname\ can also be extended to a multi-class classification setup and we conducted a 3-class classification task on the BRCA dataset.

%-------------------------------------------------------------------------
%

\subsection{Inducing Expert Concepts with LLM.}
\label{Inducing Domain Concept with LLM}
\protect \showifred{
To reduce the task difficulty and fully exploit the power of the CLIP-based pathology foundation models and human expert prior knowledge, \modelname\ decomposes a complex WSI analysis task into several patch-level subtasks - scoring related medical expert concepts. 
In this section, we detail how to induce patch-level expert concepts from human priors using a large model.
As shown in Supplementary Figure~2, in \modelname, given a specific WSI analysis task, medical literature was first collected using two search engines: Google and New Bing, with Google serving as the most powerful traditional search engine and new Bing serving as the recent search engine powered by large language models (LLMs).
Specifically, ``<key words>”, ``<key words>, journal” and ``<key words>, paper” will be searched in both engines and also Google Scholar, for instance, the ``<key words>” in lung cancer subtyping task would be ``lung adenocarcinoma” and ``lung squamous cell carcinoma”.
Then, we keep and consolidate the results of medical papers published in well-known journals such as Nature Series journals to ensure their reliability. 
Once collected, each medical literature is fed into large-language models, we utilize GPT-4 in our study, to induce instance-level expert concepts.
}

\protect \showifred{
To ensure the quality of the induced instance-level expert concepts, we adopt a three-step strategy:
First, we ask GPT-4 to summarize the pathological factors related to the classes of the target task from the input literature, together with their descriptions.
Such instance-level concepts typically correspond to potential phenotypes or clinical, pathological, and molecular characteristics that may appear in histopathology images.
Then, GPT-4 is prompted to rank the summarized factors to facilitate the final manual examination of all expert concepts conducted by the users or pathologists. 
Finally, we require GPT-4 to re-write the descriptions of the ranked pathological factors to include more visual descriptions of the tissue for fully exploiting the power of the CLIP-based pathology vision-language foundation models.
We provide a complete example of processing one medical literature related to the lung cancer subtyping task in Supplementary Figure~10. 
We feed one paper each time to GPT-4 to ensure an easier and more reliable summarization process.
With all collected medical literature being processed, we merged the summarized pathological factors from each paper, and manually deleted the repeated ones, and the factors could not be observed from histology images to form the final instance-level expert concepts groups for each target class.
Notably, as shown in Supplementary Figure~2, the generated concepts are traceable to the users as the source literature is specified in their generation process, which further ensures \modelname\ could benefit from the human expert prior knowledge in a reliable manner.
}

\protect \showifred{
When fed to \modelname, each instance-level expert concept is composed of two parts:
The first part is the above-induced text description and the second part is a learnable vector, which follows the idea of the learnable prompt representation proposed in CoOp~\cite{zhou2022learning} to improve the overall performance of the CLIP-based models (\protect\showifminor{Figure~\ref{fig:overview}e}).
}

\protect \showifred{
For the bag-level expert class prompts, \modelname\ requires GPT-4 to induce comprehensive descriptions of different target classes concerning the instance-level expert concepts induced in the previous step among each literature, then we prompt GPT-4 to merge the descriptions from all collected literature. Each bag-level expert class prompt also has two parts, similar to the instance-level expert concept (\protect\showifminor{Figure~\ref{fig:overview}e}).
All detailed automatically collected medical literature and induced instance-level expert concepts and bag-level expert class prompts can be found in our released code repository.
}
%
%-------------------------------------------------------------------------

\subsection{Learning Complementary Data-driven Concepts.}
Given a WSI $\boldsymbol{X}$ under $20\times$ magnification, we first apply the sliding window strategy to crop $\boldsymbol{X}$ into numerous non-overlapping image patches.
Then, \modelname\ extracts the instance feature from each image patch with the image encoder from \protect\showifred{the CLIP-based pathology vision-language foundation models. In this study, unless otherwise specified, we utilized QuiltNet~\cite{ikezogwo2023quilt} as the default utilized CLIP-based pathology vision-language foundation model.}
The instance feature extraction is defined as:
\begin{equation}
\boldsymbol{Z}=f_{image}\left(\boldsymbol{X}\right),
\end{equation}
where $f_{image}$ is the image encoder, and $\boldsymbol{X}$ contains $n$ cropped patches.
$\boldsymbol{Z} \in \boldsymbol{R}^{n \times d}$ is the extracted instance features, where $d$ represents the dimension of the features.
Then, we use \protect\showifred{the text encoder from the CLIP-based pathology vision-language foundation models} to obtain instance-level concept representations for each target class:
\begin{equation}
    \boldsymbol{C}_{ins}^{A}=f_{text}(\boldsymbol{T}_{ins}^{A}),
\end{equation}
where $f_{text}$ is the text encoder, $\boldsymbol{T}_{ins}^{A}$ contains $m$ instance-level concepts for target class $A$, and $\boldsymbol{C}_{ins}^{A} \in \boldsymbol{R}^{m \times d}$ is the instance-level concept embeddings for class $A$.
Specifically, $\boldsymbol{T}_{ins}^{A}$ is composed of two groups:
\begin{equation}
    \boldsymbol{T}_{ins}^{A} = \{\boldsymbol{I}_{ins}^{A}, \boldsymbol{D}_{ins}^{A}\}, 
\end{equation}
where $\boldsymbol{I}_{ins}^{A}$ is the induced instance-level expert concepts for class $A$ in the previous paragraph.
$\boldsymbol{D}_{ins}^{A}$ is a group of learnable data-driven instance-level concepts for class $A$, containing a set of purely learnable prompt representations optimized \protect\showifred{with the training data} during the training process.
The data-driven instance-level concepts $\boldsymbol{D}_{ins}^{A}$ serve as the complementary diagnostic factors to the extracted expert domain concepts $\boldsymbol{I}_{ins}^{A}$, helping to describe the whole picture of a disease.
In addition, to ensure that the learned instance-level data-driven concepts extract complementary information to the instance-level domain concepts, we define a mutual distinctive loss among them as:
\begin{equation}
    Loss_{mutual}=\sum_{\substack{i,j \in \{1,...,m\} \\ i \neq j,  cls \in \{A,B\}}}\cos\left( C_{ins, i}^{cls} \cdot C_{ins, j}^{cls}\right),
\end{equation}
where $C_{ins, i}^{cls}$ and $C_{ins, j}^{cls}$ are instance-level concept embeddings in the corresponding class.
\protect\showifred{
To avoid confusion, we emphasize that the instance-level expert concepts and the data-driven instance-level concepts in \modelname\ are quite different, and summarize their difference into the generation difference and the objective difference for easier distinguishment. 
For the generation difference: The expert concepts are generated from the human prior knowledge, as shown in Supplementary Figure~1a, they are induced by GPT-4 from the medical literature which are highly related to the target WSI analysis task and collected from the Internet. In contrast, as shown in Supplementary Figure~1b, the data-driven concepts are extracted from the WSI training dataset by \modelname\ itself and are optimized during the training process using the gradient descent algorithm.
For the objective difference: The objective of involving expert concepts is to utilize human expert prior knowledge to facilitate the overall performance of automatic WSI analysis. On the other hand, the objective of extracting data-driven concepts is to learn/extract useful knowledge for automatic diagnosis from the training data with neural networks.
Therefore, they might contain complementary knowledge beyond human pathologists, which is probably complementary to the prior knowledge contained in the expert concept, and thus benefit the overall performance of \modelname.
}

\subsection{Hierarchical Two-stage Concept-Guided Aggregation.}
We elaborately apply a hierarchical two-stage aggregation paradigm to obtain the overall bag representation under the guidance of instance-level concepts and correlations among the \protect\showifred{bag-level class prompts} and instance-level concepts in \modelname.

%
% \subsubsection{Instance Concept Guided Aggregation}
%
In the first stage, \modelname\ aggregates the extracted instance-level features $\boldsymbol{Z}$ into concept-specific bag-level features for different target classes.
For example, for class $A$, with the guidance from the instance-level concept embeddings $\boldsymbol{C}_{ins}^{A}$, the aggregation process can be formulated as:
\begin{gather}
    \boldsymbol{W}^{A}_{ins}=\operatorname{Softmax}\left(\boldsymbol{Z} \cdot {\boldsymbol{C}_{ins}^{A}}^{T}\right), \\
    \boldsymbol{H}^{A}={\boldsymbol{W}^{A}_{ins}}^{T} \cdot \boldsymbol{Z},
\end{gather}
where $\boldsymbol{W}^{A}_{ins} \in \boldsymbol{R}^{n \times m}$ is the \protect\showifred{similarity scores} between different instances and instance-level expert concepts.
With $\boldsymbol{W}^{A}_{ins}$ serves as the aggregation weights,
and $\boldsymbol{H}^{A} \in \boldsymbol{R}^{m \times d}$ aggregated as concept-specific bag-level features for class $A$, \protect\showifred{we involve multiple medical concept scoring subtasks in \modelname\ to reduce the task complexity and fully exploit the power of human prior and CLIP-based pathology foundation models.}
%

%
% \subsubsection{Bag Concept Guided Aggregation}
%
In the second stage, to obtain overall bag-level representation for class $A$, \modelname\ aggregates the concept-specific bag-level features $\boldsymbol{H}^{A}$ according to the correlations among the \protect\showifred{bag-level class prompts} and the instance-level concepts of class $A$:
\begin{gather}
    \boldsymbol{W}^{A}_{i2b}=\operatorname{Softmax}\left(\boldsymbol{C}_{ins}^{A} \cdot {\boldsymbol{C}_{bag}^{A}}^{T}\right), \\
    \boldsymbol{F}^{A}={\boldsymbol{W}^{A}_{i2b}}^{T} \cdot \boldsymbol{H}^{A} + \operatorname{mean}\left(\boldsymbol{H}^{A}\right),
\end{gather}
where $\boldsymbol{W}^{A}_{i2b} \in \boldsymbol{R}^{m \times 1}$ is the \protect\showifred{similarity scores between bag-level expert class prompt} and instance-level concepts for class $A$,
and $\boldsymbol{F}^{A} \in \boldsymbol{R}^{1 \times d}$ is the overall bag-level representation of WSI $X$ for class $A$.

Inspired by clip-adapter~\cite{gao2021clipadapter}, we also implement slide-adapters before aligning the overall bag-level representation $\boldsymbol{F}^{A}$ and the \protect\showifred{bag-level class prompt} embedding $\boldsymbol{C}_{bag}^{A}$.
Specifically, the slide-adapters serve as additional bottleneck layers to learn new features and perform residual-style feature blending with the original features aggregated from the pre-trained encoders' feature space.
In summary, the slide-adapters can be written as follows:
\begin{gather}
    {SA}_v(\boldsymbol{F}^{A}) = \operatorname{LeakyReLU}\left({\boldsymbol{F}^{A}}^T \cdot \boldsymbol{W}_1^v\right) \cdot \boldsymbol{W}_2^v, \\
    {SA}_t(\boldsymbol{C}_{bag}^{A}) = \operatorname{LeakyReLU}\left({\boldsymbol{C}_{bag}^{A}}^T \cdot \boldsymbol{W}_1^t\right) \cdot \boldsymbol{W}_2^t, \\
    {\boldsymbol{F}^{A}}^{\star} = \alpha {SA}_v({\boldsymbol{F}^{A}})^T+(1-\alpha){\boldsymbol{F}^{A}}, \\
    {\boldsymbol{C}_{bag}^{A}}^{\star} = \beta {SA}_t(\boldsymbol{C}_{bag}^{A})^T+(1-\beta) \boldsymbol{C}_{bag}^{A},
\end{gather}
where both $SA_v(\cdot)$ and $SA_t(\cdot)$ represent two layers of learnable linear transformations ${W}_1^v, {W}_2^v, {W}_1^t,$ and ${W}_2^t$ that compose the slide-adapters, with $\alpha$ and $\beta$ as adjustable hyper-parameters.
Similarly, we can obtain ${\boldsymbol{F}^{B}}^{\star}$ and ${\boldsymbol{C}_{bag}^{B}}^{\star}$ for class $B$ and any other classes. Following the CLIP method~\cite{radford2021learning}, the prediction probability can be computed as:
\begin{equation}
    p(y=A \mid \boldsymbol{X})=\frac{\exp \left(\cos \left({\boldsymbol{F}^{A}}^{\star}, {\boldsymbol{C}_{bag}^{A}}^{\star}\right) / \tau\right)}
    {\sum_{\substack{j=A}} \exp \left(\cos \left({\boldsymbol{F}^{j}}^{\star}, {\boldsymbol{C}_{bag}^{j}}^{\star}\right) / \tau\right)}.
\end{equation}
Here, $\cos\left( \cdot, \cdot\right)$ denotes the cosine similarity, and $\tau$ represents the temperature of the Softmax function.
%
%-------------------------------------------------------------------------

\subsection{Post-hoc Interpretation.}
Model interpretation and diagnosis are crucial for medical applications. \modelname\ offers post-hoc interpretation by visualizing the similarity scores between instance-level features and instance-level concepts as similarity maps.
\protect\showifred{
To generate the post-hoc interpretable similarity maps,  we visualize the similarity scores of the above patch-level concept scoring subtasks conducted on different patches of the corresponding slide. 
Highlighted regions indicate higher responses of these patches to specific medical concepts, providing detailed multi-dimensional reference information on how \modelname\ evaluates various medical factors relevant to the WSI analysis task and reaches a final diagnosis.
}

%-------------------------------------------------------------------------

\subsection{Differences from CLIP-based Models and Discussion}
\label{Difference from CLIP-based Models}
\protect\showifred{
To avoid confusion and distinguish \modelname\ from previous CLIP-based pathology foundation models, we would like to further discuss their differences.
We would like to emphasize that the objectives of the CLIP-based pathology foundation models and \modelname\ are quite different: the CLIP-based pathology foundation models aim to align path-level histology images with their corresponding text description, while \modelname\ aims to provide precise and interpretable slide-level analysis via leveraging human expert prior knowledge and complementary knowledge extracted from training data as mentioned in the above sections.
Therefore, instead of proposing a new CLIP-based pathology foundation model, \modelname\ investigates how to incorporate existing CLIP-based pathology foundation models as a basic building component to link human expert knowledge and histology images, to improve slide-level WSI analysis tasks.
}

\protect\showifred{
As shown in the left part of \protect\showifminor{Figure~\ref{fig:overview}d}, the current CLIP-based pathology foundation models can conduct patch-level classification tasks by calculating the similarity between the input histology image patch and different class prompts.
However, due to the large size of whole slide images (WSIs), it is infeasible to directly apply current CLIP-based pathology foundation models for slide-level classification, as the default input image size of current CLIP-based pathology foundation models is usually 224$\times$224.
Therefore, in previous CLIP-based pathology foundation model works (e.g., CONCH), a top-k pooling paradigm~\cite{lu2023visual} is usually adopted when dealing with slide-level classification tasks.
Specifically, as shown in the right part of \protect\showifminor{Figure~\ref{fig:overview}d}, a WSI is first tiled into numerous image patches, which could be fed into CLIP-based pathology foundation models to obtain patch-level predictions. 
Then, the slide-level prediction is calculated as the top-k pooling results of the patch-level predictions.
\modelname\ has several advantages/novelties over this top-k pooling paradigm, and could significantly improve slide-level classification tasks using the CLIP-based pathology foundation models as a basic building component.
Specifically, as mentioned in the above sections and shown in \protect\showifminor{Figure~\ref{fig:overview}d}, \modelname's advantages/novelties mainly lie in two aspects: First, \modelname\ decomposes complex WSI analysis task into subtasks and utilizes human expert prior knowledge for more accurate and interpretable prediction. Second, \modelname\ extracts complementary data-driven knowledge from the training data.
In particular, the above top-k pooling with only class prompts on patch-level could be referred to as ``one-stage aggregation”, while \modelname\ with expert and data-driven concept prompts on patch-level and class prompts on slide-level could be referred to as ``two-stage hierarchical aggregation”.
}

\protect\showifred{
Although we have different objectives from the CLIP-based pathology foundation models and their top-k pooling paradigm, to ensure a more comprehensive comparison, we conducted the following experiments in Supplementary Figure~3 to illustrate the significant advantages of \modelname.
On all the WSI analysis tasks, we observed that \modelname\ significantly improved the performance when incorporating different CLIP-based pathology foundation models, which validated the effectiveness of our model design.
}

\subsection{Data Preprocessing}
Following the default settings outlined by CLAM~\cite{lu2021data}, we initiate our pipeline by cropping the requisite image patches from each digitized slide. The process begins with the automated segmentation of tissue regions. Each WSI is loaded into memory at a downsampled resolution (32$\times$ downscale) and converted from the RGB to the HSV color space to facilitate segmentation.
To identify tissue regions (foreground), we generate a binary mask by thresholding the saturation channel of the HSV image, subsequent to applying median blurring to smooth the image edges. This step is complemented by morphological closing operations aimed at filling in small gaps and holes within the tissue regions. The approximate contours of these foreground objects are then delineated, filtered based on a predefined area threshold, and earmarked for downstream processing.
Post-segmentation, exhaustive cropping of $448 \times 448$ patches is performed within the segmented foreground contours at $20 \times$ magnification for each slide. This meticulous process ensures that the patches are representative of the histological features relevant for subsequent analyses.

\subsection{Baseline Models}
\label{sec:baseline}
Multiple Instance Learning (MIL)~\cite{maron1998framework} has been extensively investigated for WSI analysis due to its weakly supervised learning paradigm. 
Generally, previous MIL methods can be divided into two groups: (1) instance-level methods~\cite{hou2016patch, feng2017deep, campanella2019clinical, xu2019camel, kanavati2020weakly}, and (2) embedding-level methods~\cite{zhu2017deep, ilse2018attention, li2021dual, lu2021data, shao2021transmil}. 
Instance-level methods first obtain instance predictions and then aggregate them into bag predictions using either average pooling or maximum pooling. 
In contrast, embedding-level methods initially aggregate instance features into a high-level bag representation, followed by constructing a classifier based on this bag representation for bag-level prediction.
However, most existing methods exclusively learn from image data, neglecting valuable prior expert knowledge that humans utilize and consider during the diagnostic process, such as pathological and molecular factors related to the disease. 
As embedding-level methods generally possess better performance, in this study, we include seven baseline models in our experimental performance comparisons, implementation details are discussed below:
\begin{itemize}
    \item ABMIL: We followed the instructions on the ABMIL Github repository: \url{https://github.com/AMLab-Amsterdam/AttentionDeepMIL}. ABMIL addresses the MIL problem by learning the Bernoulli distribution of a bag. It introduces a neural network-based, permutation-invariant aggregation operator with an attention mechanism~\cite{ilse2018attention}.

    \item DeepAttnMISL: We followed the guidelines on the DeepAttnMISL Github repository: \url{https://github.com/uta-smile/DeepAttnMISL}. DeepAttnMISL employs both siamese MI-FCN and attention-based MIL pooling. This method efficiently learns imaging features from WSIs and aggregates WSI-level information to patient-level data~\cite{yao2020}.

    \item CLAM-SB: We followed the instructions on the CLAM Github repository: \url{https://github.com/mahmoodlab/CLAM}. CLAM utilizes attention-based learning to identify diagnostically valuable sub-regions within a slide and refines the feature space through instance-level clustering over these identified regions~\cite{lu2021data}.

    \item GTP: We followed the guidelines on the GTP Github repository: \url{https://github.com/vkola-lab/tmi2022}. GPT combines a graph-based representation of a WSI with a vision transformer to process pathology images~\cite{zheng2021deep}. 
    
    \item TransMIL: We followed the instructions on the TransMIL Github repository: \url{https://github.com/szc19990412/TransMIL}. TransMIL leverages a Transformer-based MIL approach (TransMIL) to analyze both morphological and spatial information in WSIs~\cite{shao2021transmil}.
    
    \item HIPT: We followed the instructions on the HIPT Github repository: \url{https://github.com/mahmoodlab/HIPT}. HIPT capitalizes on the inherent hierarchical structure of WSIs, employing two levels of self-supervised learning to learn high-resolution image representations for precise WSI analysis~\cite{chen2022scaling}.
    
    \item TOP: \protect \showifred{
    As the author's official implementation of TOP seems to be incomplete according to their paper's description, we also re-implemented it to ensure a comprehensive comparison.
    }
    Unlike other baselines, TOP explored integrating language prior knowledge from large language models (LLMs) and vision-language models from the natural image domain to address few-shot weakly supervised learning for WSI analysis~\cite{qu2023rise}.
    Nonetheless, their discussion primarily focuses on the low data regime and exhibits several limitations, including unreliable prior knowledge generation from LLMs, misalignment between histopathology images and vision-language models from the natural image domain, and unsatisfactory performance in a full training setup. In contrast, \modelname\ is specifically designed to overcome these issues, setting it apart from TOP.
    \protect\showifred{
    Specifically, as shown in Supplementary Table~1, has advantages over the limitations of TOP in three aspects - summarizing traceable expert concepts from related medical literature, finetuning visual-textual feature spaces using slide adapters, and extracting complementary knowledge from training WSIs with learnable data-driven concepts.
    In addition, we have also compared the average AUC and ACC among all five tasks of \modelname\ and TOP's original implementation among different CLIP-based pathology models in 
    Supplementary Figure~10 alone.
    we observe that \modelname\ consistently outperforms TOP's original implementation, which validates the advantages of \modelname\ over TOP's limitations.
    }
\end{itemize}

\subsection{Training Details}  
All experiments were conducted on a workstation equipped with eight NVIDIA RTX 3090 GPUs.
\protect\showifred{
Unless otherwise specified, we employed the image encoder ViT-B/32 and text encoder GPT/77 of QuiltNet~\cite{ikezogwo2023quilt} in \modelname\ as the feature extractors for both histopathology images and textual concepts in this study. 
}
The length of learnable parts was set to 16 tokens for both instance-level and bag-level expert concepts.
For the number of instance-level concepts, we utilized 26 instance-level concepts for each target class, while the number of learned instance-level concepts was tuned from $\{2, 4, 6, 8, 10, 12\}$ for each target class. 
During the training, we fixed all weights of the CLIP-based pathology foundation models and only learned the learnable part of the instance-level expert concepts bag-level expert class prompts, and the purely learnable data-driven concepts.
\protect\showifred{Note that the above-learned parts are all input for the CLIP-based pathology foundation models.}
For model optimization, we employed the SGD optimizer and a batch size of 2. The learning rate was set to 0.0001 for all datasets. 
For evaluation metrics, the area under the curve (AUC) of receiver operating characteristic, and the accuracy (ACC) were adopted. 
We utilized patient-level five-fold cross-validation for all experiments and reported the results of all models in the form of mean and standard deviation.

\subsection{Ethics Approval and Consent to Participate}
\protect\showifred{
All datasets and other materials employed in this study are previously published and publicly accessible with approved protocol and participants's informed consent.
No new human research participants are involved in this study.
}
%-------------------------------------------------------------------------

\section{Data Availability}\label{data_avail_sec}
All datasets and other materials employed in this study are previously published and publicly accessible.
The TCGA datasets was acquired from the Genomic Data Commons Data Portal at \url{https://portal.gdc.cancer.gov/}.
Source data and the medical literature including the extracted human expert concepts for this study are provided alongside this paper's released code repository \protect \showifred{(\url{https://github.com/HKU-MedAI/ConcepPath})}.

\section{Code Availability}\label{code_avail_sec}
The \modelname\ source code is available in Github (\url{https://github.com/HKU-MedAI/ConcepPath}). 
We also uploaded all scripts and materials to reproduce all the analyses on the same website.
\protect\showifminor{
A tutorial Colab notebook, including the trained weights of the studied clinical tasks, is also provided.
}

\section{Acknowledgements}\label{acknowledgement_sec}

This work was partially supported by the Research Grants Council of the Hong Kong SAR, China (Project No. 27206123 and T45-401/22-N) and the Hong Kong Innovation and Technology Fund (Project No. ITS/274/22).

\section{Author contributions}\label{author_contributions_sec}
LY conceived and supervised the study.
MY supervised the study and provided expert opinions on pathology concepts and data interpretation for model development.
YJ reviewed and provided expert opinions for this study.
WZ, ZG, and YF implemented the framework and performed all data analysis.
WZ, and ZG wrote the manuscript with inputs from all authors.
All authors reviewed and approved the final paper.

\section{Competing Interests}
The authors declare no competing interests.

%%%%%%%%%%%%%%%%%%%%%%%%%%%%%%%%%%%%%%%%%%%%%%%%%%%%%%%%%%%%%%%%%%%%%%%%%%%%%%%%%%%%%%%%%%%%%%%%%%%%%%%%%%%%%%%%%%
% \bibliography{sn-bibliography}

%%%%%%%%%%%%%%%%%%%%%%%%%%%%%%%%%%%%%%%%%%%%%%%%%%%%%%%%%%%%%%%%%%%%%%%%%%%%%%%%%%%%%%%%%%%%%%%%%%%%%%%%%%%%%%%%%%%

\begin{figure*}[!ht]
\centering
\includegraphics[width=0.95\linewidth]{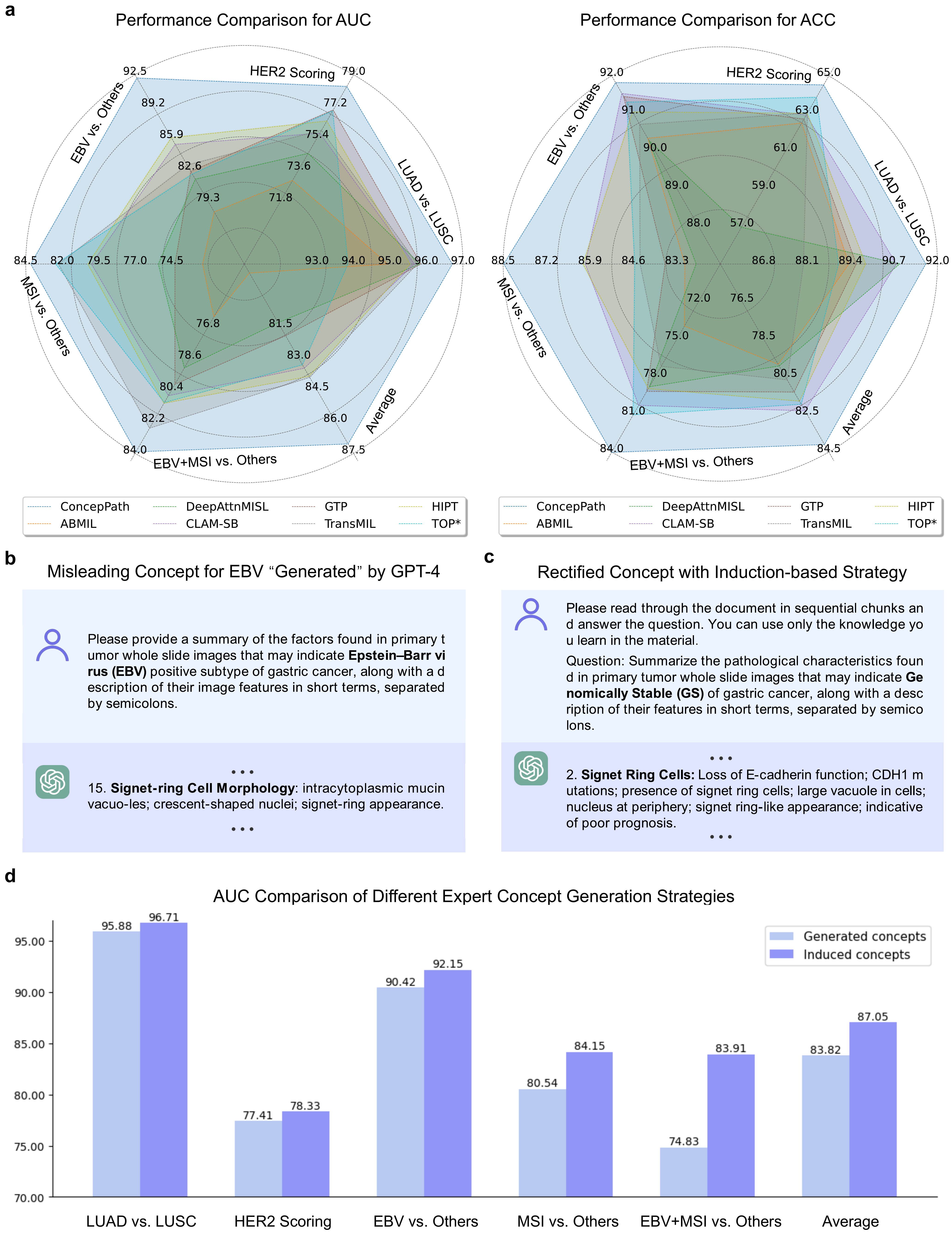}
    \caption{
        \textbf{Performance and expert concept generation comparison of \modelname.}
        \protect\showifred{
        \textbf{a,}}
        Radar charts depicting the average AUC(Left) and ACC(Right) for the five WSI analysis tasks in the five-fold cross-validation experiment conducted on NSCLC, BRCA, and STAD datasets. 
        \protect\showifred{``Average” denotes the average performance among all five tasks. ``TOP*” represents the higher performance in the author's implementation and our implementation of TOP. }
        \modelname\ demonstrated more accurate predictions on all five tasks since it successfully incorporates human expert prior knowledge and data-driven knowledge learned from the training data.
        \textbf{b,c,} A misleading concept generated by directly querying GPT-4 (denoted as ``Generated”),  
    }
\end{figure*}

\begin{figure}[!t]
    \ContinuedFloat
    \captionsetup{labelformat=empty}  % 取消 "Figure X"
    \caption{ 
        which our induction-based method (denoted as ``Induced”) successfully rectifies for the gastric immunotherapy-sensitive subtyping task. Specifically, the concept ``signet-ring cells” was found in the category of Epstein–Barr virus (EBV) positive subtype in the generated concepts; however, this morphology is more commonly linked to the Genomically Stable (GS) subtype, where mutations in CDH1 and RHO genes play a pivotal role.
        \protect\showifred{\textbf{d,}}
        % %
        A histogram representing the AUC comparison of different expert concept generation strategies. The y-axis is the AUC($\%$) and the x-axis is the WSI analysis tasks and their average performance. ``Induced” concepts demonstrated better performance among all tasks, especially for the more challenging immunotherapy-sensitive subtyping tasks on the STAD dataset, highlighting the importance of inducing concepts from professional materials for complex WSI analysis tasks.
        \vspace{20cm}
    }
\label{fig:performance}
\end{figure}

\begin{figure*}[!b]
\centering
\includegraphics[width=0.95\linewidth]{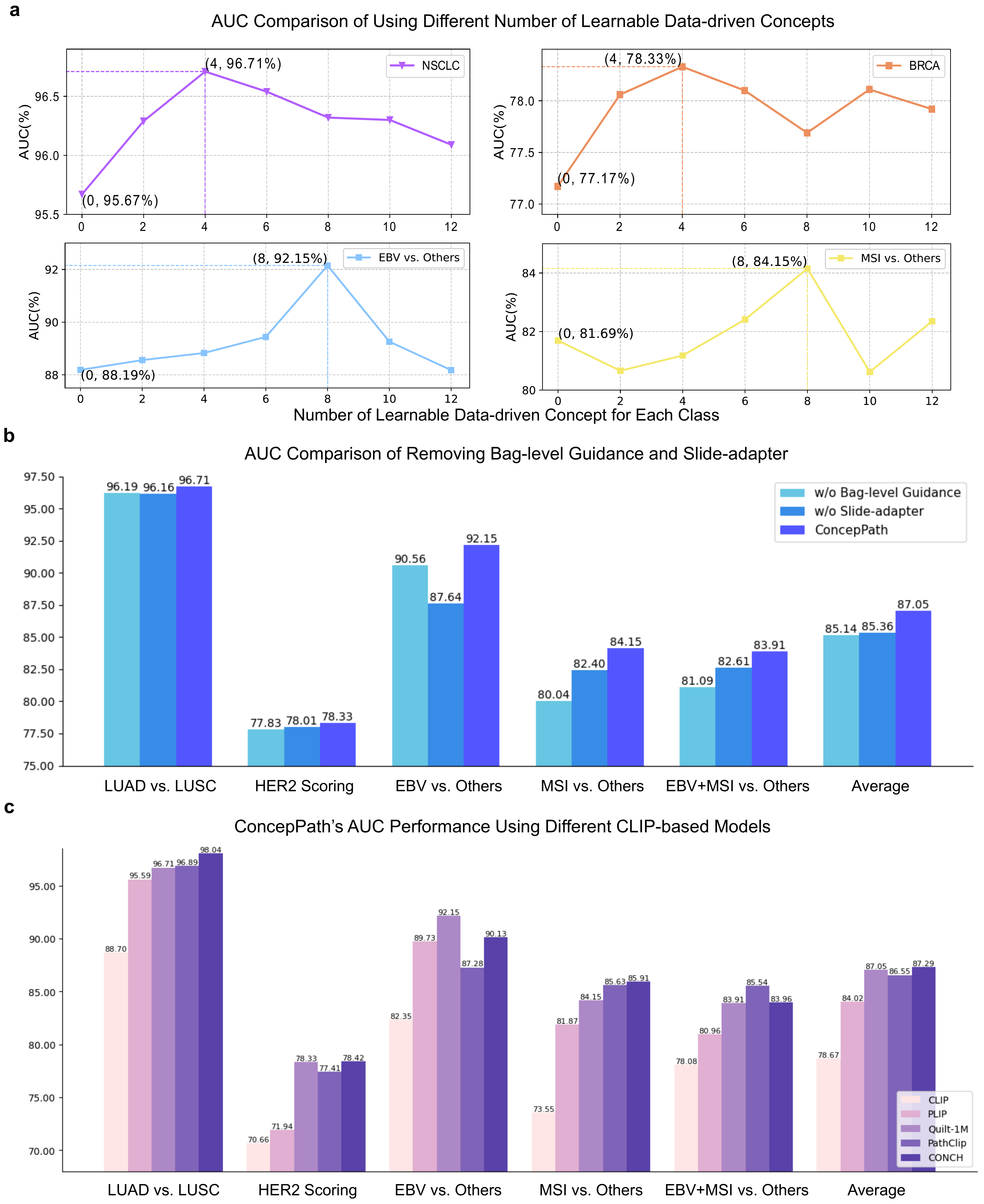}
    \caption{
    \textbf{Investigation of proposed components in \modelname.}
    \textbf{a,} Line plots for investigating new data-driven concept learning, the y-axis is the AUC($\%$) and the x-axis is the number of learned concepts used for each class, where 0 means only using human expert concepts. Integrating data-driven knowledge learned from training data improves overall performance, and for more challenging and inadequately researched diagnostic tasks, a greater number of learned concepts are required. The performance decline when the number of learned concepts was large suggests a potential trade-off between prior expert knowledge and learned knowledge.
    \protect\showifred{\textbf{b,}}
    A histogram representing investigations on second stage bag-level concept guided aggregation and slide-adapters, the y-axis is the AUC($\%$), and ``w/o Bag-level guidance” refers to using average pooling aggregation. Both modules contributed to the improved performance.
    \protect\showifred{\textbf{c,}}
    \protect\showifred{A histogram representing the comparison of using different CLIP-based vision-language models as ConcepPath’s basic component for aligning}
    }
\end{figure*}

\begin{figure}[!t]
    \ContinuedFloat
    \captionsetup{labelformat=empty}  % 取消 "Figure X"
    \caption{
    \protect\showifred{for aligning histopathology images and concept knowledge, and the y-axis is the AUC($\%$). }
    Obvious performance increase can be observed by using pathology vision-language models, and \modelname\ could benefit from more accurate alignment if the pathology vision-language were trained on larger datasets.
    }
    
    \label{fig:ablation}
\end{figure}

\begin{figure}[!ht]
    \centering
    \includegraphics[width=0.98\textwidth]{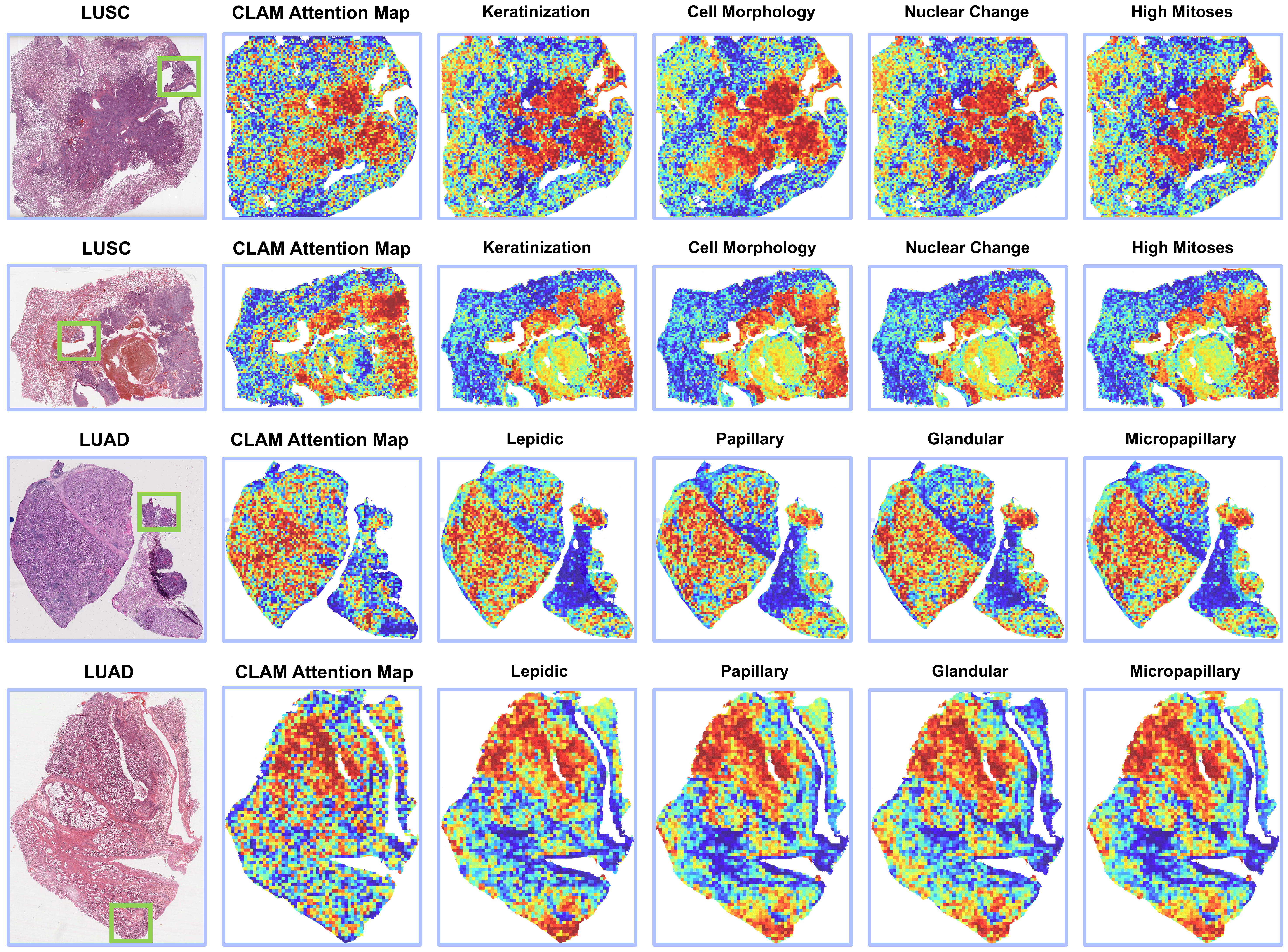}
    
    \caption{
            \protect\showifminor{
                \textbf{Visualizations of \modelname\ and baseline method.}
                    Instance-level expert concept similarity maps.
                    The slides are accurately identified as the lung squamous cell carcinoma (LUSC) and lung adenocarcinoma (LUAD) subtype, respectively. 
                    In comparison to the CLAM attention map, the similarity maps of various instance-level concepts provide a more precise focus on the tumor in the green box highlighted area.
            }
                    \vspace{7.0cm}
    }
    
    \label{supp_sim}
\end{figure}

\clearpage

\end{document}